\documentclass[extended-abstract]{ccn}

\title{QualiaNet: An Experience-Before-Inference Network}

\author{Paul Linton \\
Columbia University \\
\email{paul.linton@columbia.edu}}
\usepackage{hyperref}
\hypersetup{
    colorlinks=true,
    urlcolor=blue
}

\begin{document}

\maketitle

\begin{abstract}
Human 3D vision involves two distinct stages: an Experience Module, where stereo depth is extracted relative to fixation, and an Inference Module, where this experience is interpreted to estimate 3D scene properties. Paradoxically, although stereo vision does not provide us with absolute distance information, it nonetheless affects our inferences about distance. We propose the Inference Module exploits a natural scene statistic: near scenes produce vivid disparity gradients, while far scenes appear comparatively flat. QualiaNet implements this two-stage architecture computationally: disparity maps simulating human stereo experience are passed to a CNN trained to estimate distance. The network can recover distance from disparity gradients alone, validating this approach.
\end{abstract}

\section{Introduction}

3D vision from multiple viewpoints is a key challenge: Tsao \& Tsao (2022), Linsley et al. (2025), O’Connell et al. (2025), Lee et al. (2026), Bonnen et al. (2026).

One dimension that hasn’t been explored is ‘consciousness’ or ‘qualia’: the ‘subjective visual experience’ associated with 3D vision. This paper is inspired by the experience of human stereo vision. When we look at a car vs a picture of a car, our \textbf{inferences} about the car’s 3D shape can be the same, even though our \textbf{experience} of the car’s 3D shape will be vastly different.

\begin{figure}[h]
    \centering
    \includegraphics[width=1\linewidth]{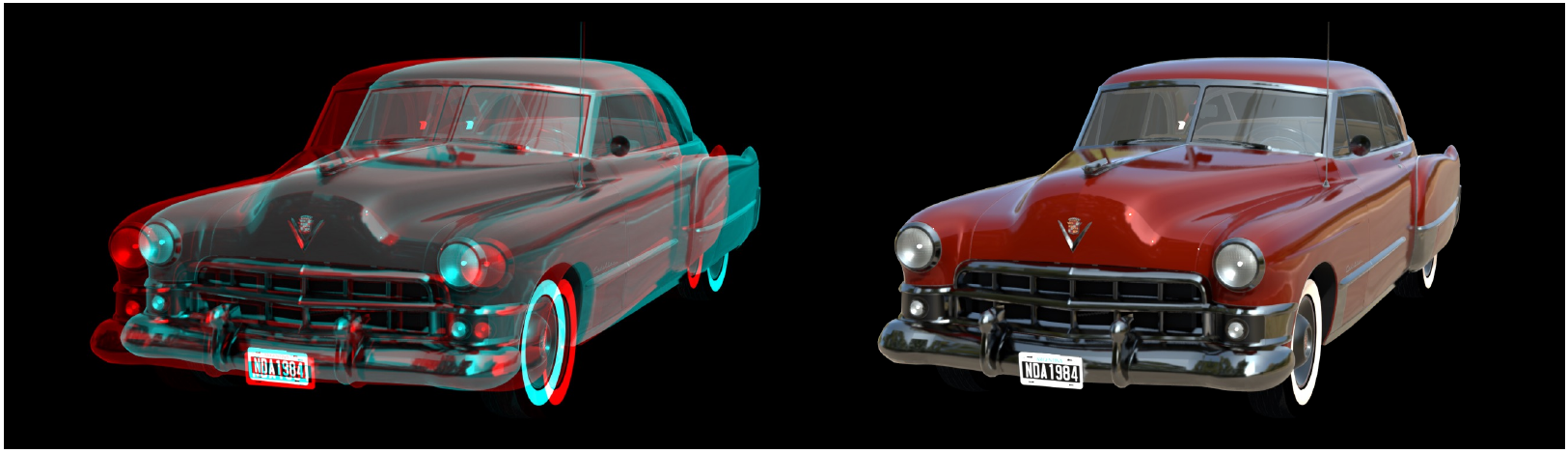}
    \caption{Stereo car vs picture of a car.}
\end{figure}

To accommodate this fundamental fact about human 3D vision, Linton (2025, 2023, 2017) argues that human vision involves two distinct stages: 

\textbf{1. Experience Module}: First, depth structure is extracted from stereo vision and experienced, explaining why the car and the picture of the car can lead to different 3D experiences. This is low-level and hard coded.

\textbf{2. Inference Module}: Second, the experienced depth structure from stereo vision is interpreted, explaining why the car and the picture of the car can lead to the same 3D inference. This is learned during development. 

\section{Experience Module (`Qualia')}

The depth structure we extract and experience from stereo vision in the Experience Module is surprisingly impoverished. It appears to simply reflect the disparity gradient (angular difference of points between the two eyes) relative to fixation (Linton, 2024a; more accurately, something very close to this: Linton, 2023). 

In this paper, we simulate this by taking a monocular depth map from Unity, setting the fixation point (zero disparity) to the center of the image (the central jug), and calculating the angular disparity from fixation of all the points in the image for different viewing distances.

\begin{figure}[h]
    \centering
    \includegraphics[width=0.8\linewidth]{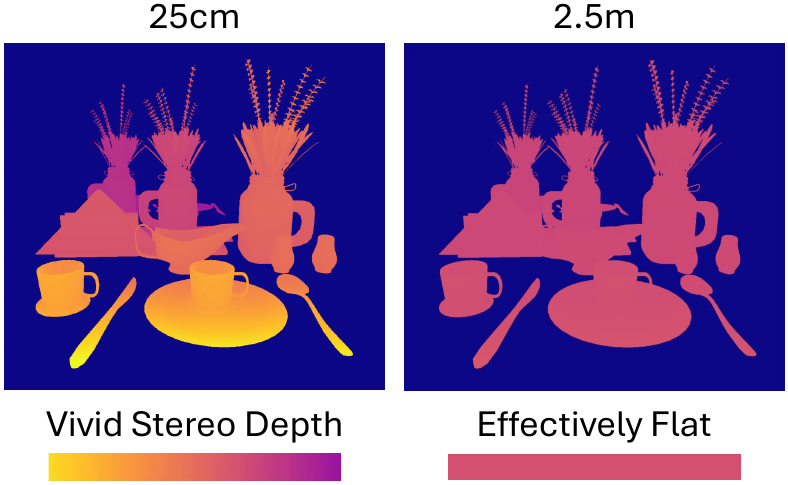}
    \caption{Disparity gradients (the size of the angular offset of points relative to fixation) simulated for 25cm (left) and 2.5m (right) viewing distances. See Figure 3.}
\end{figure}

\begin{figure*}[h]
    \centering
    \includegraphics[width=0.975\linewidth]{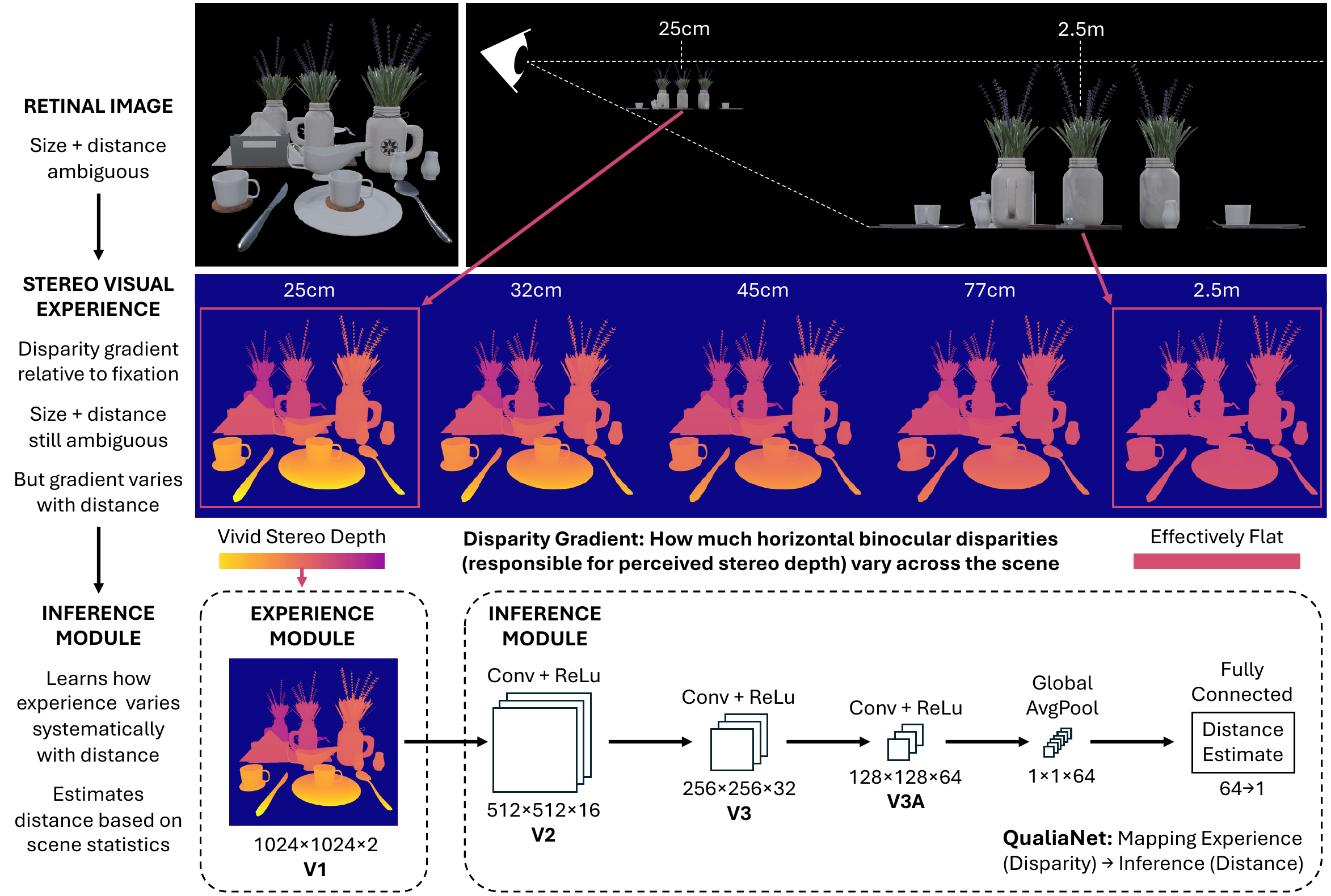}
    \caption{QualiaNet from Retinal Image $\rightarrow$ Experience Module (disparity gradient) $\rightarrow$ Inference Module (distance).}
\end{figure*}

Two things become immediately apparent: 

\textbf{1. Visual Scale}: The scale (size and distance) of the scene is ambiguous. The central jug is the same shade of orange in the two scenes, even though it is 25cm away in one scene and 2.5m away in the other.

\textbf{2. 3D Shape}: The disparity gradient is large for near scenes (producing vivid stereo depth) and small for far scenes (effectively flat), meaning our visual experience of 3D shape is distorted with viewing distance. 

\section{Inference Module (Post-`Qualia')}

The Inference Module infers 3D scene properties - Visual Scale and 3D Shape - from our distorted visual experience of stereo depth (the disparity gradient).

The surprising thing is that even though stereo vision provides no absolute distance information, it has a powerful effect on visual scale. Helmholtz (1857) demonstrated the effect of stereo vision on visual scale, showing that if we artificially increase the separation between the eyes (using mirrors) the world seems miniature.

The explanation in Linton (2021a, 2023) is that the Inference Module learns how stereo depth varies systematically with viewing distance: near scenes produce large disparity gradients, whereas far scenes appear comparatively flat. Scale is estimated not by triangulating absolute distance, but by interpreting disparity-defined visual experience in light of natural scene statistics. Effectively using one deficit of stereo vision (the distortion of 3D shape with distance) to compensate for the other (the absence of absolute distance). An illusion presented at the 2025 Vision Sciences Society suggests that this is how human vision works (Linton, 2024b).

\section{QualiaNet}

QualiaNet implements this two-stage architecture (Experience Module $\rightarrow$ Inference Module) computationally. 

\textbf{1. Experience Module}: The Experience Module is simulated by taking a scene in Unity, scaling it to  different distances (central jug: 25cm to 2.5m), and calculating the angular disparity from fixation (central jug) for all the points in the image. All the monocular cues in the image are fixed, leaving the CNN with only disparity gradients to rely on. This first stage is hypothesized to correspond to feedforward processing in V1 (Linton, 2021b).

\textbf{2. Inference Module}: This 1024 x 1024 disparity map relative to fixation (plus a 1024 x 1024 mask to exclude background pixels) is then fed into a CNN that is trained to estimate the absolute distance of fixation using pairs of disparity maps + ground truth absolute distances.

The network is loosely inspired by the dorsal visual stream. The input takes up 56° of the visual field, and the CNN receptive field sizes increase from 0.59°/11px ($\approx$V2), to 2.74°/51px ($\approx$V3), to 9.2°/171px ($\approx$V3A).

\textbf{Training}: The network is trained on 600 disparity map + absolute distances, using 100 randomly selected 1/d distances between 1/25cm and 1/2.5m, applied to (1) scene, (2) scene minus near objects, (3) scene minus far objects, (4-6) horizontally flipped versions of (1-3).

\textbf{Testing + Results}: Tested on 200 examples of a new version of the scene with objects rearranged. Accurately recovers distance (R² = 0.97, RMSE = 0.08m).

\begin{figure}[h]
    \centering
    \includegraphics[width=0.65\linewidth]{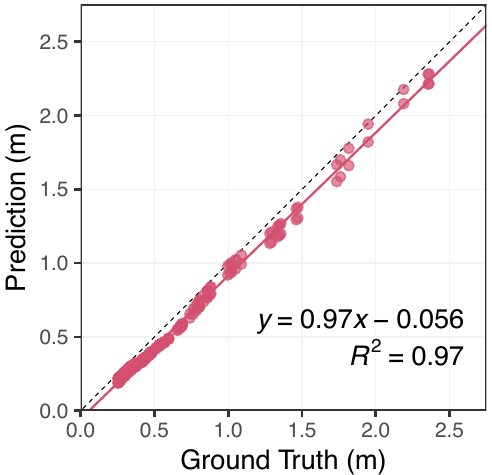}
    \caption{QualiaNet accurately recovers absolute distance on a new scene using disparity gradients alone.}
\end{figure}

\newpage
\noindent\textbf{Project Page:} 
\href{https://QualiaNet.GitHub.io/}{\textbf{QualiaNet.github.io}} 

\section{Acknowledgments}

This research project and related results were made possible by the support of the NOMIS Foundation. This research was conducted in Nikolaus Kriegeskorte's Visual Inference Lab at Columbia University's Zuckerman Mind Brain Behavior Institute as a NOMIS Foundation Fellow at the Italian Academy for Advanced Studies in America. I thank the NOMIS Foundation ('New Theory of Visual Experience' grant to PL), the Italian Academy for Advanced Studies, Columbia University, and the Presidential Scholars in Society and Neuroscience (PSSN), Columbia University, for their support.

\nocite{Linton2017}
\nocite{Linton2021a}
\nocite{Linton2021b}
\nocite{Linton2023}
\nocite{Linton2024a}
\nocite{Linton2024b}
\nocite{Linton2025}
\nocite{Helmholtz1857}
\nocite{Tsao&Tsao2022}
\nocite{Linsley2025}
\nocite{OConnell2025}
\nocite{Lee2026}
\nocite{Bonnen2026}

\printbibliography

@Book{Linton2017,
  author =	 {Paul Linton},
  title =	 {The Perception and Cognition of Visual Space},
  publisher =	 {Palgrave Macmillan},
  year =	 2017,
  address =	 {Cham, CH}
}

@Article{Linton2021a,
  author =	 {Paul Linton},
  title =	 {Does Vergence Affect Perceived Size?},
  journal =	 {Vision},
  year =	 2021,
  volume =	 5,
  pages =	 {33}
}

@Article{Linton2021b,
  author =	 {Paul Linton},
  title =	 {V1 as an egocentric cognitive map},
  journal =	 {Neuroscience of Consciousness},
  year =	 2021,
  volume =	 2,
  pages =	 {niab017}
}

@Article{Linton2023,
  author =	 {Paul Linton},
  title =	 {Minimal theory of 3D vision: new approach to visual scale and visual shape},
  journal =	 {Phil Trans Royal Soc B},
  year =	 2023,
  volume =	 378,
  pages =	 {20210455}
}

@Article{Linton2024a,
  author =	 {Paul Linton},
  title =	 {Linton Stereo Illusion},
  journal =	 {ArXiv},
  year =	 2024,
  pages =	 {2408.00770}
}

@Article{Linton2024b,
  author =	 {Paul Linton},
  title =	 {Visual Scale is Governed by Horizontal Disparities: Linton Scale Illusion},
  journal =	 {PsyArXiv},
  year =	 2024,
  pages =	 {ywbj4\_v1}
}

@Article{Linton2025,
  author =	 {Paul Linton},
  title =	 {2025 {Marr} Medal Lecture: Experience Before Inference},
  journal =	 {Applied Vision Association},
  year =	 2025,
}

@Article{Helmholtz1857,
  author =	 {Helmholtz, Hermann},
  title =	 {Das Telestereoskop [The telestereoscope]},
  journal =	 {Annalen der Physik und Chemie},
  year =	 1858,
  volume =	 101,
  pages =	 {494}
}

@Article{Tsao&Tsao2022,
  author =	 {Thomas Tsao and Doris Y. Tsao},
  title =	 {A topological solution to object segmentation and tracking},
  journal =	 {PNAS},
  year =	 2022,
  volume =	 119,
  pages =	 {e2204248119}
}

@InProceedings{Linsley2025,
  author =	 {Drew Linsley and Peisen Zhou and Alekh Karkada Ashok and Akash Nagaraj and Gaurav Gaonkar and Francis E Lewis and Zygmunt Pizlo and Thomas Serre},
  title =	 {The 3D-PC: a benchmark for visual perspective taking in humans and machines},
  booktitle = {International Conference on Learning Representations ({ICLR})},
  year      = {2025},
}

@Article{OConnell2025,
  author =	 {Thomas P. O'Connell and Tyler Bonnen and Yoni Friedman and Ayush Tewari and Vincent Sitzmann and Joshua B. Tenenbaum and Nancy Kanwisher},
  title =	 {Approximating Human-Level 3D Visual Inferences With Deep Neural Networks},
  journal =	 {Open Mind},
  year =	 2025,
  volume =	 9,
  pages =	 {305}
}

@InProceedings{Lee2026,
  author    = {Wanhee Lee and Klemen Kotar and Rahul Mysore Venkatesh and Jared Watrous and Honglin Chen and Khai Loong Aw and Daniel L. K. Yamins},
  title     = {Unified {3D} Scene Understanding Through Physical World Modeling},
  booktitle = {International Conference on Learning Representations (ICLR)},
  year      = {2026},
}

@Article{Bonnen2026,
  author =	 {Tyler Bonnen and Jitendra Malik and Angjoo Kanazawa},
  title =	 {Human-level 3D shape perception emerges from multi-view learning},
  journal =	 {ArXiv},
  year =	 2026,
  pages =	 {2602.17650}
}
\end{document}